# Top-Down Feedback for Crowd Counting Convolutional Neural Network

**Deepak Babu Sam** and **R. Venkatesh Babu**
Video Analytics Lab, Indian Institute of Science,
Bangalore 560012, INDIA
{deepaksam, venky}@iisc.ac.in

## Abstract

Counting people in dense crowds is a demanding task even for humans. This is primarily due to the large variability in appearance of people. Often people are only seen as a bunch of blobs. Occlusions, pose variations and background clutter further compound the difficulty. In this scenario, identifying a person requires larger spatial context and semantics of the scene. But the current state-of-the-art CNN regressors for crowd counting are feedforward and use only limited spatial context to detect people. They look for local crowd patterns to regress the crowd density map, resulting in false predictions. Hence, we propose top-down feedback to correct the initial prediction of the CNN. Our architecture consists of a bottom-up CNN along with a separate top-down CNN to generate feedback. The bottom-up network, which regresses the crowd density map, has two columns of CNN with different receptive fields. Features from various layers of the bottom-up CNN are fed to the top-down network. The feedback, thus generated, is applied on the lower layers of the bottom-up network in the form of multiplicative gating. This masking weighs activations of the bottom-up network at spatial as well as feature levels to correct the density prediction. We evaluate the performance of our model on all major crowd datasets and show the effectiveness of top-down feedback.

## Introduction

From busy streets to massive gatherings, crowd count often serves as a metric of great practical importance. Increasingly it is infeasible to employ humans to do crowd analysis, boosting the case for building automated systems. Crowd counting in Computer Vision sense, is an instance recognition task that estimates the number of people in an image. However, in extremely dense crowds, people occupy only few pixels, making it difficult to individually detect and count people. Many times people are seen as bunch of blobs, which requires some level of reasoning about the scene to conclude the presence of humans. Severe occlusion, pose changes, view-point variations and illumination conditions further complicate the task. Some typical crowd scenes are displayed in Figure 1.

Hence crowd counting problem is often relaxed to a regression task which is nowadays been done by Convolu-tional Neural Networks (CNN). Given an image, counting CNNs are trained to regress spatial crowd density, i.e. number of people per unit area. They are forced to learn a hierarchy of filters specific to crowd features as opposed to individual human features. For example, these CNNs model edges to detect head-shoulder pair as they appear in a crowded scene, rather than facial features like eyes or nose. As a result, many crowd like patterns in the image created by leaves of trees, buildings, cluttered backgrounds etc. are misclassified as people. This causes lot of false predictions as shown in Figure 1. Some high level context information would have indicated that these are irrelevant patterns. This is a general problem with any feedforward bottom-up systems, where the low-level feature detectors do not have enough context information to decide on the input. Useful information might be lost in the initial layers (say, by spatial max-pooling etc.) of a neural network, which could be needed for the correct prediction. Ideally, once high-level context of the image is known, the system has to evaluate it and arrive at the final decision.

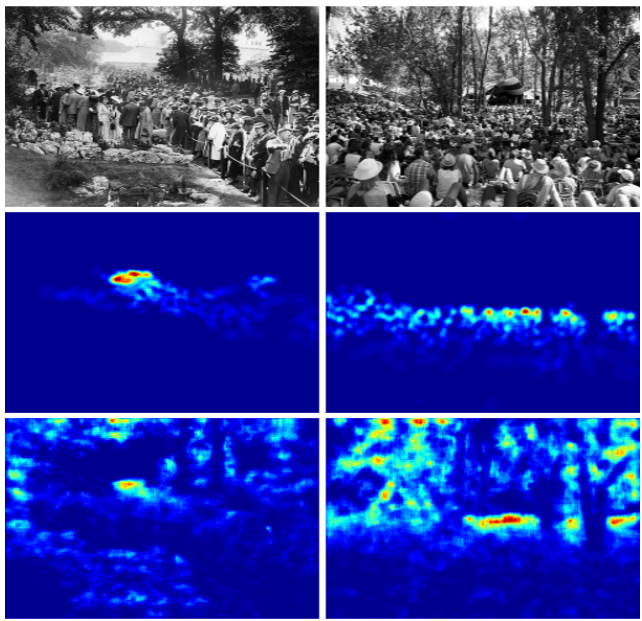

Figure 1: Density maps predicted by a typical CNN regressor (bottom row) have a lot of false detections. Many crowd like patterns are identified as humans. Top row displays the input scene and middle row holds the ground truth density map.

One way to address these problems is to look at how humans do crowd counting. Whenever there is difficulty in identifying people, say at extremely dense crowd, people use high level scene understanding to judge whether the crowd like blobs are actually humans. Also, there is ample evidence from neuroscience research that brain has complex feedback networks (Lamme, Super, and Spekreijse 1998; Gilbert and Sigman 2007; Piëch et al. 2013). Information flows in both directions; high level cortical areas can influence low level feature detectors. In this paper, we try to mimic some aspects of the top-down feedback mechanism to solve crowd counting task.

Our primary aim is to use high-level context information about the scene to correct false density predictions of the counting CNN. To that end, we construct a regular CNN density regressor as the bottom-up network and a separate top-down CNN to utilize scene context. Top-down information comes in the form of feature maps output by higher layers of the bottom-up CNN. This information is used by the top-down network to generate feedback. The feedback here acts as a correcting signal to the lower layers of the bottom-up network. In our work, feedback is applied in the form of multiplicative gating to the low-level feature activations of the bottom-up network. The density map is generated once again after applying the feedback to the bottom-up CNN. In this way, the lower layers of the CNN regressor receive high-level context information in the form gating. The feedback gates the lower layer feature activations of the bottom-up CNN to correct the density prediction.

In summary, the major contributions of this paper are:

- A generic architecture to deliver top-down information in the form of feedback to the bottom-up network.

- A crowd counting system that uses top-down feedback framework to correct its density predictions.

## Related Work

Many early works in crowd counting from images rely on head detections. Appearance based hand-crafted features are used in (Wu and Nevatia 2005; Wang and Wang 2011) to train detectors. Since these systems build on head features, they perform suboptimally with dense images, where majority of those appearance features are irrelevant. Hence, models leveraging crowd features become more popular. Information from various sources are fused in (Idrees et al. 2013) to do crowd counting. They use HOG based head detectors, SIFT interest point vectors and Fourier transform features to regress the count.

CNN based regressors outperform all other methods that use hand-crafted features. Zhang et al. (Zhang et al. 2015) train a counting CNN by alternatively backpropagating crowd density loss and crowd count loss. To get cross-scene applicability, they fine-tune the CNN only with training images similar to target scene. Since this similarity is found using density and perspective information, their model is limited by the availability of such data. In (Wang et al. 2015), a deep CNN is employed to regress directly the crowd count instead of a density map. However, direct count predicting models fail to learn good features, leading to high error rates. In order to account the large variation in appearance of crowd, Onoro-Rubio et al. (Onoro-Rubio and López-Sastre 2016) feed images at different scales to separate CNNs. Each of the CNN has the same architecture and is trained on images of single scale. Their outputs are fused with fully connected layers to get the final density prediction. But the choice of number of scales significantly affects the performance of their model and varies across datasets. Another way to tackle scale variation is with multiple columns of CNN with different receptive fields. In (Boominathan, Kruthiventi, and Babu 2016), the density prediction of a deep CNN based on VGG-16 network is fused with that of a shallow CNN. The deep CNN looks for crowd at higher scale, while the shallow CNN is specialized for very dense crowds. Zhang et al. (Zhang et al. 2016) use three columns of CNN having different filter sizes to capture crowd scenes at multiple scales. The features of these CNN columns are fused together to regress the density map. To impose more clear specialization between multiple CNN columns of varied architectures, a differential training scheme is used by (Sam, Surya, and Babu 2017). A classifier is also trained to route the given test image patches to their corresponding expert columns. Even though such multi-column models work quite well, they incur lot of false detections on crowd like irrelevant patterns. This arises because of the absence of high-level context information to the lower layers of the CNN regressor.

In contrast, such top-down influences exist in the human brain (Gilbert and Sigman 2007; Piëch et al. 2013). These top-down modulations are evoked by a complex network of horizontal and feedback connections (Lamme, Super, and Spekreijse 1998). They help the lower visual areas to attain attentional selectivity, so that only relevant information is combined and spurious responses are removed (Desimone 1998; Beck and Kastner 2009). Some works in Computer Vision try to incorporate some of these aspects. Many works (Gatta, Romero, and van de Veijer 2014; Shrivastava and Gupta 2016; Ranjan and Black 2016; Li, Hariharan, and Malik 2016) use a series of networks to iteratively perform their tasks of interests. Initial output by the first network is fed to the next network along with context information. Feature maps of the previous network is concatenated with its output to supply top-down information to the next network in series. Main drawback of this approach is that each stage requires a separate network to be trained. Instead of iteratively improving predictions with multiple similar network, (Pinheiro et al. 2016; Shrivastava et al. 2016) have a separate top-down network which takes features from different layers of the bottom-up network as context information. Note that the top-down network generates the final output and no feedback is given back to the bottom-up network. Hence the top-down net-work also learns to do the task of interest using high-level context, rather than providing feedback to the bottom-up network.

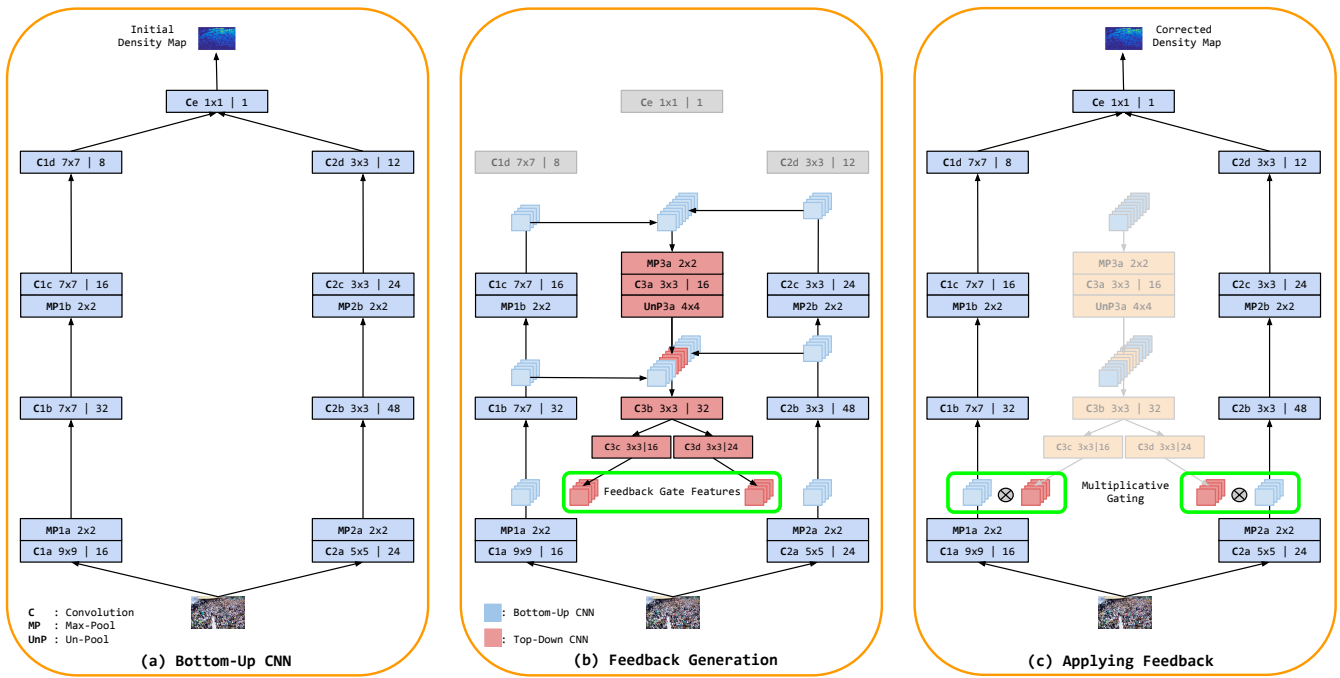

Figure 2: Architecture of Top-down feedback CNN. (a) displays bottom-up CNN, (b) depicts the feedback generation by the top-down CNN and (c) shows how the bottom-up CNN re-evaluates its prediction using the gate features. Best viewed in colour.

On the other hand, in our architecture, the top-down network learns to drive feedback to the lower layers of the bottom-up network. Here, the form of feedback is also important. Works like (Stollenga et al. 2014; Wang et al. 2014; Cao et al. 2015) use multiplicative mask on feature maps to suppress unwanted activations. We also apply feedback in the form of multiplicative gating. The final prediction is output by the bottom-up network using the feedback gated activations.

## Our Approach
### Feedback as a Correcting Signal

As discussed in Introduction, the majority of the crowd counting systems rely on a CNN to regress the crowd density map. However, all these regressing models do not individually detect and count people. They learn to look for crowd features (how the heads of a bunch of people appear) as opposed to individual person features (eyes, nose, body parts etc.). Consequently, many crowd like patterns (trees, cluttered backgrounds, etc.) are detected as people. Also, the density prediction at various regions could be wrong due to occlusions or background interferences. Many of these problems could be solved if high-level context information is available to the density regressor. Hence, our approach is to use top-down feedback as a correcting signal to re-evaluate the density prediction of the CNN. A separate top-down network learns to correlate the high-level context of the scene with the low-level responses of the CNN regressor. It generates masks, which weigh spatially all feature activations of the lower layers. This suppresses spurious detections and pass legitimate responses to generate the corrected density map.

### Top-Down Feedback CNN

Our model has a bottom-up CNN for density prediction and a separate top-down CNN for feedback generation. A two column CNN forms the bottom-up network. Feature maps from final layers of the CNN columns are fused with a $1 \times 1$ filter to obtain the density map. This CNN regressor is similar to the architecture introduced in (Zhang et al. 2016). We use it because of the simplicity and performance it offers. The CNN columns basically vary in filter sizes and hence the receptive fields. These are designed to capture crowd at different scales. The first network has large initial filter size of $9 \times 9$ and can attend patterns that span large receptive fields like big faces etc. The other column with initial filter size $5 \times 5$ is meant for dense crowds. Both are shallow networks with only four convolutional layers and two pooling layers. Figure 2(a) shows the bottom-up network.

The top-down CNN runs down parallel to the bottom-up network. It consumes high-level feature maps from the bottom-up network to generate feedback. Generally, the feature maps are taken from the layers immediately before pooling layers of the bottom-up CNN. This is to ensure that the top-down network has access to any relevant information lost in spatial pooling. As depicted in Figure 2(b), the feature maps from the convolutional layers C1c and C2c of the bottom-up CNN are concatenated and pooled. Max-pooling MP3a is done so that convolutional layer C3a receives larger spatial context about the crowd scene. Since the bottom-up network is pretrained, the features of the last layers C1d and C2d resemble more like density maps. Those features are not given to the top-down network as they do not add useful context information. The features maps of C3a are unpooled (values are repeated) to make it same size as the corresponding features of the bottom-up network. The resultant feature maps are again concatenated to that of the bottom-up network. These are again operated by two convolutional layers to generate the feedback gate features. The activation function for C3c and C3d are chosen to be sigmoid so that the values are in 0-1 range. All other convolutional layers use ReLU non-linearity. Hence, the gate features can strongly damp spurious activations or allow legitimate responses to pass. It has the same spatial and feature dimensions as that of the output of the first pool layers MP1a and MP2a.

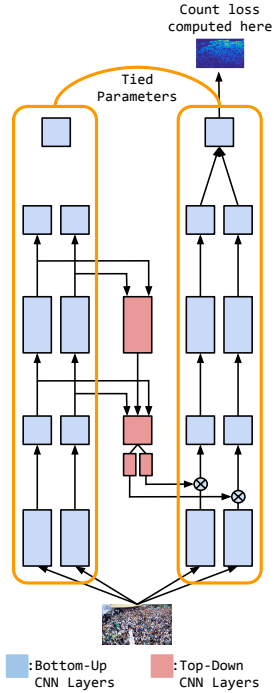

Figure 3: The unrolled computation graph used for training top-down network. The top-down CNN uses features of the bottom-up CNN to apply feedback so that the bottom-up network can be re-evaluated. So, the bottom-up CNN appears twice in the computation graph and the loss is calculated on the corrected prediction. Only parameters of the top-down CNN are updated. Best viewed in colour.

The feedback is applied by element-wise multiplication of the gate feature maps with that of the bottom-up CNN. It provides different spatial gating for each of the feature maps of the first convolutional layers C1a and C2a. So, the top-down network influences the bottom-up CNN at spatial as well as individual feature level. Moreover, it could selectively control information flow through the two columns of the bottom-up CNN. Depending on the scale of the crowd, one CNN column can be given prominence over the other. The $5 \times 5$ CNN column which performs better for dense crowds, needs to have more effect in the final prediction of such scenes. Note that it is not useful to apply feedback gating directly on the input image. Doing so will mask many regions of the image and destroy relevant context required for higher layers (see Analysis section).

Density prediction happens in two steps as shown in Figure 2(b,c); first the image is passed through the bottom-up network. Its features are used to compute the gate feedback maps. The gate features are then element-wise multiplied with feature maps of the bottom-up network to generate the corrected density map. The model is trained also in two stages. Initially, the bottom-up CNN is trained and its parameters are fixed. This is followed by training of the feedback network.

### Training of Bottom-Up CNN

The bottom-up CNN is initially trained alone to regress crowd density maps. Both columns of the bottom-up network are individually pretrained. This ensures learning of better features and makes later finetuning more effective. The $l_2$ distance between the predicted density map and ground truth is used as the loss to train the CNN regressor. If $D_{X_i}(x; \Theta)$ stands for the output of a CNN regressor with parameters $\Theta$, the $l_2$ loss function is given by

$$L_{l_2}(\Theta) = \frac{1}{2N} \sum_{i=1}^{N} \|D_{X_i}(x; \Theta) - D_{X_i}^{GT}(x)\|_2^2, \quad (1)$$

where $N$ is the number of training samples. Here, ground truth density map is $D_{X_i}^{GT}(x)$ for the input image $X_i$. Standard Stochastic Gradient Descent (SGD) algorithm is applied on the parameters $\Theta$ to optimize $L_{l_2}$. The $l_2$ loss function implicitly captures the count error between the predicted and ground truth count. Minimizing $L_{l_2}$ reduces count error.

Many methods have been proposed for generating density maps from ground truth head annotations. Most popular way is to simply blur each head annotation with a Gaussian kernel normalized to sum to one. Summing the density map gives the crowd count. This kind of ground truth makes regression easier for the CNN, as it no longer needs to get the exact point of head annotation right. The spread of the Gaussian kernel is chosen depending on the dataset. Training is done with patches which have 1/4th size of the original image. 9 patches are cropped at different locations from every image to augment the data as in (Zhang et al. 2016).

### Training of Top-Down CNN

After the bottom-up network is trained, its parameters are fixed. The top-down network is trained by backpropagating the loss incurred by the estimated density after applying feedback. The unrolled computation graph is shown in figure 3. While training, only the parameters of the top-down network are updated. The parameters are updated so as to reduce the count error of the final prediction. The top-down

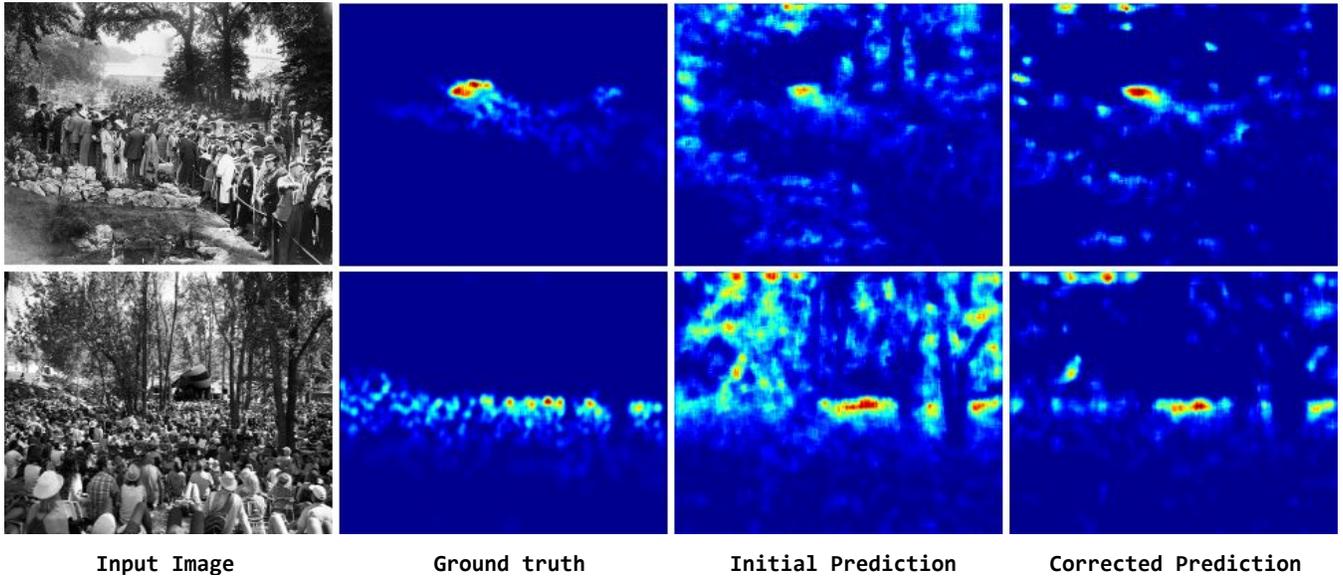

Figure 4: Sample predictions of TDF-CNN on images of Part_A of Shanghaitech dataset (Zhang et al. 2016).

CNN, thus learns to use context information to gate the activations of the bottom-up network and correct the density prediction.

The performance of any counting CNN is measured using count error. The crowd count of an image $X_i$ is computed from its density prediction $D_{X_i}$ as $C_{X_i} = \sum_x D_{X_i}(x; \Theta)$. Let its actual count be $C_{X_i}^{GT} = \sum_x D_{X_i}^{GT}(x)$. Then, count loss is the squared difference between predicted and true count,

$$L_C(\Theta) = \frac{\lambda}{2N} \sum_{i=1}^{N} (C_{X_i} - C_{X_i}^{GT})^2, \qquad (2)$$

where $\lambda$ is a constant multiplier to check the magnitude of the loss. We use $l_2$-loss between the density maps to train the bottom-up network, since it accounts for spatial distribution of the density and helps learn better features. But, reducing the count error between the estimated and ground truth map is our primary objective. Hence, it is beneficial to use count loss rather than $L_{l_2}$(eq: 1) loss for training top-down network. In this way, top-down CNN receives complementary information to improve the overall performance. Experimentally, we find that using $L_{l_2}$(eq: 1) loss for top-down training does not result in much improvement. Note that training top-down network with count loss, still learns good features as the bottom-up network is already trained.

We also add $l_1$ regularizer to the loss function to impose sparse activations for the feedback gate features. This aids the top-down network to train effectively and allows only relevant features to be active. So, the final loss function for the top-down CNN is,

$$L_{TD}(\Theta) = \frac{\lambda}{2N} \sum_{i=1}^{N} (C_{X_i} - C_{X_i}^{GT})^2 + \frac{\mu}{N} \sum_{i=1}^{N} G_{X_i}, \qquad (3)$$

where the scalar $G_{X_i}$ is the sum of all feedback gate features generated for image $X_i$ and $\mu$ is a regularization constant. The values of the regularizer constants are chosen empirically. In all our experiments, these regularizers are fixed as $\lambda = 10^{-2}$ and $\mu = 10^{-3}$. The loss $L_{TD}$ is used to back-propagate the top-down CNN till the validation accuracy plateaus.

## Experiments

### Evaluation Scheme

Our model, TDF-CNN is evaluated on four major crowd counting datasets. During testing, the input image passes through the TDF-CNN to generate feedback gate features. The feedback features are then applied on the bottom-up CNN to predict the final density map. The predicted density maps are 1/4th size of the image because of the two pooling layers.

We use two metrics to benchmark TDF-CNN, namely Mean Absolute Error (MAE) and Mean squared error (MSE). MAE is defined as

$$\text{MAE} = \frac{1}{N} \sum_{i=1}^{N} |C_{X_i} - C_{X_i}^{GT}|, \qquad (4)$$

where $C_{X_i}$ is the count estimated by TDF-CNN, $C_{X_i}^{GT}$ is the actual count and $N$ is the total number of images in the test set. MAE is representative of the accuracy of the model. Further, to account the variance of estimation, MSE is computed as,

$$\text{MSE} = \sqrt{\frac{1}{N} \sum_{i=1}^{N} (C_{X_i} - C_{X_i}^{GT})^2}. \qquad (5)$$

In other words, MSE indicates the robustness of the count prediction. To have fair comparison, we also analyse models using the number of parameters they employ.

## Shanghaitech dataset

The Shanghaitech crowd counting dataset, which consist of 1198 annotated images, is introduced by (Zhang et al. 2016). It has two parts named Part_A and Part_B. Part_A includes 482 images collected from the Internet which are mostly dense with the number of people varying from 33 to 3139. On the other hand, 716 images in Part_B are taken from busy streets in Shanghai. They are less dense with crowd counts between 9 to 578. For Part A, 300 images are used for training and the rest 182 images for testing. Similarly, 400 images of Part_B are for training and 316 for testing. The ground truth head annotation are convolved with a Gaussian kernel to obtain ground truth density maps. Here, we use geometry-adaptive kernels method (Zhang et al. 2016) to adapt the variance of the Gaussian depending on the crowd density.

Table 1 reports performance of TDF-CNN along with other models. It is seen that TDF-CNN outperforms all other models by a significant margin both in terms of MAE and MSE. Moreover, the number of parameters for TDF-CNN is less than all other models. This emphasizes effectiveness of top-down feedback in correcting density predictions. Figure 4 shows some density predictions of TDF-CNN along with the initial predictions. Comparing predictions without feedback, it is observed that many false detections are removed in the corrected maps.

|        | Part_A |       | Part_B |      | Prm  |
|--------|--------|-------|--------|------|------|
| Method | MAE    | MSE   | MAE    | MSE  |      |
| LBP+RR | 303.2  | 371.0 | 59.1   | 81.7 | -    |
| Zhang et al. | 181.8 | 277.7 | 32.0 | 49.8 | 0.62 |
| MCNN (2016) | 110.2 | 173.2 | 26.4 | 41.3 | 0.15 |
| TDF-CNN | **97.5** | **145.1** | **20.7** | **32.8** | **0.13** |

Table 1: Comparison of TDF-CNN to other methods on Part_A and Part_B of Shanghaitech dataset (Zhang et al. 2016). Our model performs better on all metrics. Prm stands for number of model parameters in millions. LBP+RR refers to a model that uses Local Binary Pattern and Ridge Regression for estimating crowd count (Zhang et al. 2016).

## UCF_CC_50 dataset

UCF_CC_50 (Idrees et al. 2013) is a small dataset of 50 annotated crowd scenes. These are highly dense crowds with counts between 94 and 4543. The small size of the dataset and large variation in crowd count makes this dataset quite challenging. Since there is no separate test set, 5-fold cross-validation is performed for evaluation of our model (Idrees et al. 2013; Zhang et al. 2015; Boominathan, Kruthiventi, and Babu 2016; Zhang et al. 2016). The ground truth is generated with a Gaussian kernel of fixed variance as in (Zhang et al. 2015; Boominathan, Kruthiventi, and Babu 2016; Zhang et al. 2016; Onoro-Rubio and López-Sastre 2016). For this dataset, we use $3 \times 3$ filters for the last convolutional layer (Ce in Figure 2) of TDF-CNN.

It is evident from Table 2 that TDF-CNN performs better than all other models except Hydra2s (Onoro-Rubio and López-Sastre 2016). The difference between the two models in terms of MAE is 21.1. But note that, TDF-CNN has roughly 93% less number of parameters than Hydra2s network. This indicates that our model with top-down feedback performs competitively with quite few parameters.

| Method | MAE | MSE | Params |
|--------|-----|-----|--------|
| Lempitsky et al. (2010) | 493.4 | 487.1 | - |
| Idrees et al. (2013) | 419.5 | 487.1 | - |
| Zhang et al. (2015) | 467.0 | 498.5 | 0.62M |
| CrowdNet (2016) | 452.5 | - | 14.72M |
| MCNN (2016) | 377.6 | 509.1 | 0.15M |
| Hydra2s (2016) | **333.7** | **425.3** | 1.82M |
| TDF-CNN | 354.7 | 491.4 | **0.13M** |

Table 2: Benchmarking of TDF-CNN on UCF_CC_50 dataset (Idrees et al. 2013). TDF-CNN performs competitively with fewer model parameters.

## WorldExpo'10 dataset

The WorldExpo'10 dataset (Zhang et al. 2015) contains 1132 video sequences captured with 108 surveillance cameras in Shanghai 2010 WorldExpo. It has 3980 frames, out of which 3380 are used for training. Test set includes five different video sequence with 120 frames each. The crowds are relatively sparse compared with other dataset. There are only 50 people per image on an average. Region of interest (ROI) is provided for both training and test scenes. In addition, the authors also provide perspective maps for all scenes. TDF-CNN is trained and tested with ROI as done in (Zhang et al. 2015; 2016). While training, error is backpropagated only for areas in the ROI. Similarly, only ROI regions are evaluated for testing. MAE is computed for each of the 5 test scenes and averaged.

Table 3 lists the performance of all methods. Despite the dataset being relatively sparse in crowd density, TDF-CNN is able to offer better MAE in three scenes as well as in average terms. This further underlines the need for top-down feedback in crowd counting.

## Analysis
### Effectiveness of Feedback

In this section, the effectiveness of top-down feedback in crowd counting is demonstrated with ablations. First, we study the performance improvement offered by top-down feedback. To that end, prediction accuracy of the bottom-up CNN is compared to that with top-down feedback in Table 4. The bottom-up CNN trained on Part_A of the Shanghaitech dataset, gives an MAE of 147.4. Interestingly, adding top-down feedback to the bottom-up network decreases the MAE to 97.5. Thus, in this case, top-down feedback is able to correct the counting error by a significant factor.

As described earlier (section Training of Top-Down CNN), the top-down CNN is trained with count loss $L_C$ (eq: 2). This ensures that the top-down network learns complementary information to correct the prediction of the bottom-up CNN. For fair comparison, we train the bottom-up CNN with count loss. The pretrained CNN columns in Figure 2(a) are fine-tuned with the final $1 \times 1$ filter using count loss. Again, from Table 4, this network performs inferior to the TDF-CNN. This clearly indicates that the performance gain observed for TDF-CNN is due to the feedback mechanism and not because of using count loss alone. Note that, fine-tuning with count loss distorts to some extent the spatial quality of the predicted density maps, which does not happen with TDF-CNN. Further, to make parameters roughly the same as TDF-CNN, we train MCNN (Zhang et al. 2016) with count loss. MCNN has three shallow CNN columns which are pretrained. Their outputs are fused with a $1 \times 1$ filter and fine-tuned by back-propagating the count loss. Still, TDF-CNN has lower MAE.

| Method | Scene1 | Scene2 | Scene3 | Scene4 | Scene5 | Average MAE | Params |
|---|---|---|---|---|---|---|---|
| LBP+RR (Zhang et al. 2016) | 13.6 | 59.8 | 37.1 | 21.8 | 23.4 | 31.0 | - |
| Zhang et al. (Zhang et al. 2015) | 9.8 | **14.1** | 14.3 | 22.2 | 3.7 | 12.9 | 0.62M |
| MCNN (Zhang et al. 2016) | 3.4 | 20.6 | 12.9 | **13.0** | 8.1 | 11.6 | 0.15M |
| TDF-CNN | **2.7** | 23.4 | **10.7** | 17.6 | **3.3** | **11.5** | **0.13M** |

Table 3: MAEs computed for 5 test scenes of WorldExpo'10 dataset (Zhang et al. 2015). Our top-down feedback model has better MAE for 3 scenes and delivers lower average MAE.

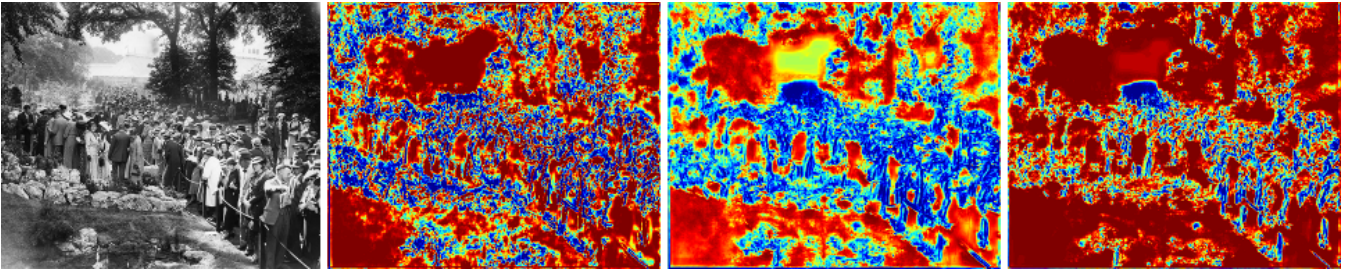

Figure 5: Some of the feedback gate maps for the input image shown in first column. Red to blue colour scale maps to 0-1 range.

The bottom-up CNN that we use, has two CNN columns with different field-of-view. Such a design allows the top-down network to not only gate features of the individual column, but also be selective about the CNN columns themselves. We show in Table 4 that the top-down feedback results in considerable performance gain even without a multi-column architecture for the bottom-up network. For this experiment, the CNN column with initial filter size $9 \times 9$ alone is used as the bottom-up network. The top-down network remains same as in Figure 2(b), except for layer C3d, which is removed. The MAE for this network with top-down feedback is 21.4% less than that without feedback. This means that the top-down framework is generic and can be applied on a variety of bottom-up networks.

In order to shed light on how the gate feedback features work, we display some of the gate feedback features in Figure 5. The images show that the gate feature maps indeed act as masks. Majority of the values of the gates are near 0. This is like selective gating, where only relevant information is allowed to pass. Note that there are many gate feature maps, only some are shown. They could be complementary also, i.e. a region blocked in one feature map may be open in another. Activations of spatial regions are not fully suppressed in all feature maps. Instead, they weigh differently across feature maps to supply relevant detections to higher layers. This is precisely the reason why we do not apply the feedback directly on the input. Gating on the input blanks out many regions of the image that might provide crucial context cues for higher layers. Also, it is interesting that some gate maps offer rough localisations of the crowd. This demonstrates that the top-down network actually learns to disambiguate people in the scene to correct the bottom-up prediction.

| Method | MAE |
|---|---|
| Bottom-up CNN | 147.4 |
| Bottom-up CNN fine-tuned with count loss $L_C$ | 116.2 |
| MCNN (2016) fine-tuned with count loss $L_C$ | 116.7 |
| Only $9 \times 9$ CNN column | 158.5 |
| TDF-CNN with only $9 \times 9$ column | 124.6 |
| TDF-CNN with $9 \times 9$ & $5 \times 5$ columns | **97.5** |

Table 4: Comparison of models with and without top-down feedback. The lower MAE delivered by models with feedback is indicative of its effectiveness.

## Conclusion

Typical crowd counting CNNs are trained to look for crowd patterns, instead of accounting every person and count. As a result, many crowd like patterns are detected as people leading to massive false predictions. In this paper, we propose top-down feedback, which carries high-level scene context to correct spurious detections. Our architecture consists of a bottom-up CNN, which has connections to another top-down CNN. The top-down CNN generates feedback in the form of gating to the lower level activations of the bottom-up CNN. This selectively passes legitimate activations and damps spurious responses. We show that such a feedback model delivers better or competitive results on all major crowd counting datasets. Further, we demonstrate the effectiveness of top-down feedback with ablations.

## Acknowledgements

This work was supported by Science and Engineering Research Board (SERB), Department of Science and Technology (DST), Government of India (Proj No. SB/S3/EECE/0127/2015).